%% file: neurips_2024.tex
\documentclass{article}

\pdfoutput=1
     \PassOptionsToPackage{numbers, compress}{natbib}
\usepackage[preprint]{neurips_2024}

\usepackage{amsmath}
\usepackage[utf8]{inputenc} % allow utf-8 input
\usepackage[T1]{fontenc}    % use 8-bit T1 fonts
\usepackage{url}            % simple URL typesetting
\usepackage{booktabs}       % professional-quality tables
\usepackage{amsfonts}       % blackboard math symbols
\usepackage{nicefrac}       % compact symbols for 1/2, etc.
\usepackage{microtype}      % microtypography
\usepackage{xcolor}         % colors
\usepackage{eucal}
\usepackage{accents,graphicx}
\usepackage{bm}
\usepackage{pifont}
\usepackage{enumitem}
\usepackage{url}
\usepackage{multirow}
\usepackage{float}
\usepackage{svg}
\usepackage{bbding}
\usepackage{utfsym}

\usepackage[colorlinks,linkcolor=red,filecolor=blue,citecolor=blue,urlcolor=blue]{hyperref}
\usepackage{cleveref}
\usepackage{array}
\usepackage{geometry}
\title{AV-DiT: Efficient Audio-Visual Diffusion Transformer for Joint Audio and Video Generation}

\author{Kai Wang$^{1, \clubsuit}$, Shijian Deng$^{2,\spadesuit}$, Jing Shi$^{3, \dagger}$, Dimitrios Hatzinakos$^{1, \heartsuit}$, Yapeng Tian$^{2, \diamondsuit}$ \\
$^1$University of Toronto, $^2$University of Texas at Dallas, $^3$Adobe Research\\
$^\clubsuit$\texttt{kaikai.wang@mail.utoronto.ca}, $^\spadesuit$\texttt{shijian.deng@utdallas.edu}\\ 
$^\dagger$\texttt{jingshi@adobe.com},
$^\heartsuit$\texttt{dimitris@comm.utoronto.ca},
$^\diamondsuit$\texttt{yapeng.tian@utdallas.edu}
}

\begin{document}

\maketitle

\input{sec/0_abstract}
\input{sec/1_introduction}

\input{sec/2_related_work}

\input{sec/3_method}
\input{sec/4_experiments}
\input{sec/5_conclusion}

\bibliographystyle{unsrt}
\bibliography{neurips_2024}

%%%%%%%%%%%%%%%%%%%%%%%%%%%%%%%%%%%%%%%%%%%%%%%%%%%%%%%%%%%%
%\input{sec/checklist}
\newpage
\input{sec/appendix}

%%%%%%%%%%%%%%%%%%%%%%%%%%%%%%%%%%%%%%%%%%%%%%%%%%%%%%%%%%%%

\end{document}

%% file: sec/0_abstract.tex
\begin{abstract}
Recent Diffusion Transformers (DiTs) have shown impressive capabilities in generating high-quality single-modality content, including images, videos, and audio. However, it is still under-explored whether the transformer-based diffuser can efficiently denoise the Gaussian noises towards superb multimodal content creation. To bridge this gap, we introduce AV-DiT, a novel and efficient audio-visual diffusion transformer designed to generate high-quality, realistic videos with both visual and audio tracks. To minimize model complexity and computational costs, AV-DiT utilizes a shared DiT backbone pre-trained on image-only data, with only lightweight, newly inserted adapters being trainable. This shared backbone facilitates both audio and video generation.
Specifically, the video branch incorporates a trainable temporal attention layer into a frozen pre-trained DiT block for temporal consistency. Additionally, a small number of trainable parameters adapt the image-based DiT block for audio generation. An extra shared DiT block, equipped with lightweight parameters, facilitates feature interaction between audio and visual modalities, ensuring alignment. Extensive experiments on the AIST++ and Landscape datasets demonstrate that AV-DiT achieves state-of-the-art performance in joint audio-visual generation with significantly fewer tunable parameters. Furthermore, our results highlight that a single shared image generative backbone with modality-specific adaptations is sufficient for constructing a joint audio-video generator. \textit{Our source code and pre-trained models will be released.}

\begin{figure*}[ht]
\centering
\vspace{-4.0em}
\includegraphics[scale=0.32]{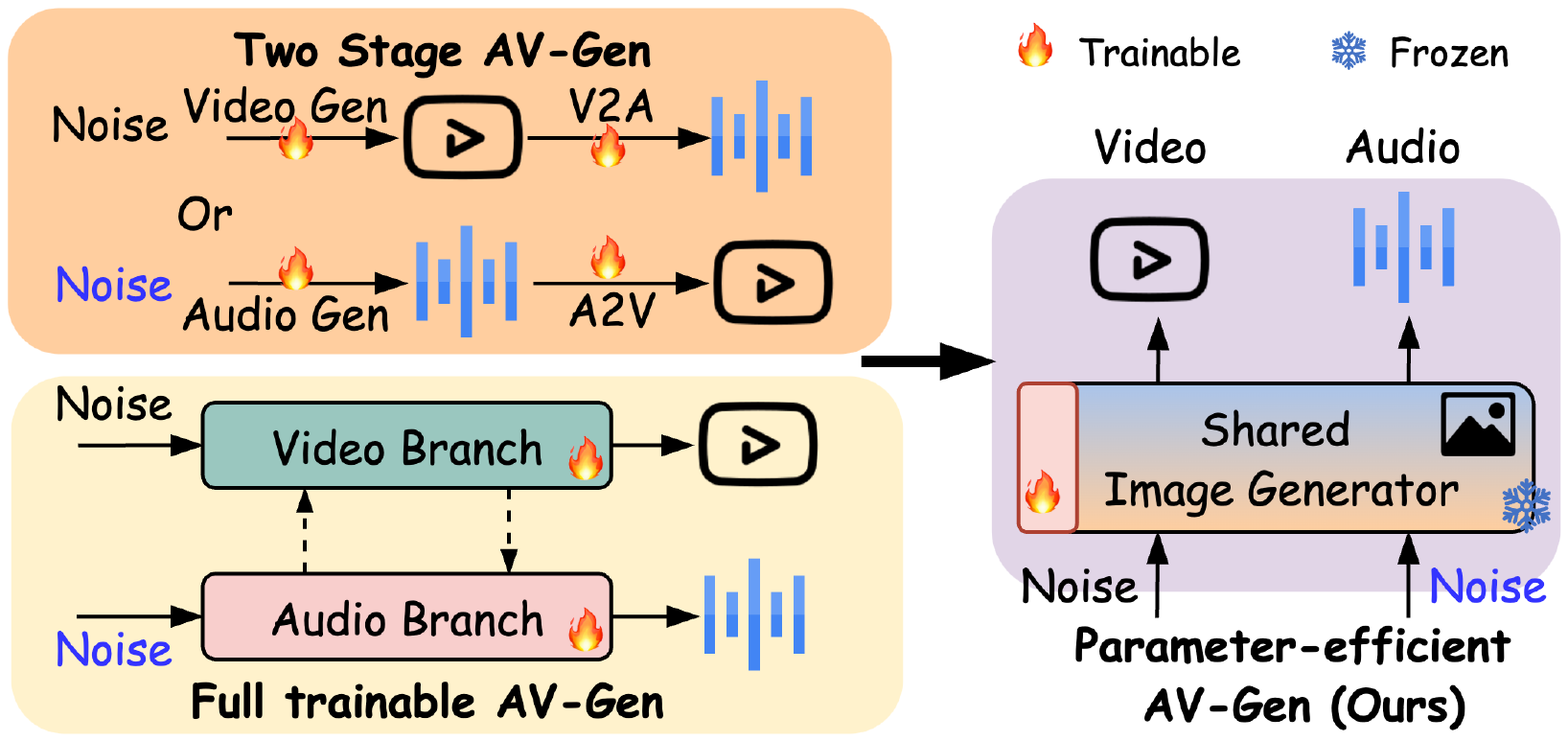}
\vspace{-4em}
\caption{Comparison of our Audio-Video Generator (AV-Gen) with existing methods.}

\label{figure 1}
\vspace{-2em}
\end{figure*}

\end{abstract}

%% file: sec/1_introduction.tex
\section{Introduction}

In recent years, diffusion models~\cite{ho2020denoising} have emerged as powerful generative technologies, significantly advancing AI-generated content creation, including images ~\cite{rombach2022high, ramesh2022hierarchical, saharia2022photorealistic}, videos~\cite{ho2022imagen, singer2022make, chen2023seine, bar2024lumiere, wang2023lavie}, and audio~\cite{liu2023audioldm, ghosal2023text, huang2023make}. However, most existing research focuses on generating single-modality content, neglecting the inherent multimodality of real-world perception and content. For instance, most video diffusion models only generate silent videos, lacking the auditory component crucial for a complete viewing experience. Our work aims to address this gap by generating high-quality videos that people can watch and listen to as shown in Figure \ref{figure 1}. 

For generating both audio and video content, one naive method is to adopt a two-stage generation that first creates silent videos via video diffusion models and then produces the corresponding audio based on the generated video. Instead, recent works like MM-Diffusion~\cite{ruan2023mm} and Seeing and Hearing~\cite{xing2024seeing} propose to simultaneously generate video and audio by convolutional U-Net based diffusion models. With Sora~\cite{videoworldsimulators2024} achieving an impressive performance in creating high-fidelity long video (without audio), diffusion transformers (DiTs)~\cite{peebles2023scalable} gradually present incredible performance in generating high-quality content by replacing the commonly used U-Net backbone with a scalable transformer. 

However, most transformer-based diffusion models are restricted to generating a single modality and their effectiveness on joint audio and video generation is still under-explored. Besides, training such diffusion transformer models on multimodal generation from scratch requires massive pairs of multimodal data and unexpected computing resources, which is not achievable to perform the implementation in the academic laboratory. Benefiting from parameter-efficient fine-tuning (PEFT) technologies~\cite{houlsby2019parameter}, the pre-trained vision foundation models can be adopted as a shared backbone for solving audio-visual understanding tasks~\cite{lin2023vision, duan2024cross} due to their excellent generalization ability and knowledge transferability. Hence, an interesting question arises: \textit{Is it possible to adapt the shared pre-trained image generative backbone equipped with modality-specific adaptations for constructing an efficient joint audio-video generator?} Our findings obtain an affirmative response to this question. 

In this paper, we propose a novel and efficient Audio-Visual Diffusion Transformer, namely AV-DiT, for simultaneously generating high-quality and realistic audio and video with minimal computation cost. Our AV-DiT utilizes a shared DiT backbone pre-trained solely on ImageNet as well as lightweight trainable layers (e.g. LoRA and adapters) to extend the image generation into joint audio and video generation. Compared with MM-Diffusion introducing an extra super-resolution module and involving full parameter updates, our end-to-end AV-DiT only trains the newly inserted layers while maintaining the pre-trained DiT backbone frozen, greatly reducing the fewer tunable parameters and computing memory. To achieve the cross-domain adaptation from the image into audio-video modality, our AV-DiT effectively tackles three challenges via the designed adaptations: \textbf{1.} Introducing the temporal consistency into frozen DiT via inserted temporal adapters for video generation; \textbf{2.} Alleviating the domain gap between image and audio by frozen DiT equipping with LoRA and adapter layers for audio generation; \textbf{3.} Enabling frozen DiT to learn the correlation between audio and video with LoRA layers for multimodal alignment. The main contributions of this work can be summarized as follows:
\begin{itemize}
  \item [1.] 
  
   We propose AV-DiT, the first multimodal diffusion transformer architecture for joint audio and video generation by leveraging off-the-shelf frozen DiT pre-trained on image-only data and minimal trainable adapters.   
   
  \item [2.]

   Our AV-DiT adapts the shared frozen pre-trained DiT for video generation by introducing temporal consistency, for audio generation by mitigating the domain gap, and for audio-visual alignment by multimodal interaction, demonstrating that the joint audio-video generation can benefit from the pre-trained image generator.

  \item [3.]
  
  Extensive experiments on the AIST++ and Landscape datasets show that our AV-DiT achieves competitive or even better performance than state-of-the-art methods in generating high-quality and realistic video and audio while involving fewer trainable parameters. 
\end{itemize}

%% file: sec/2_related_work.tex
\section{Related Work}

\textbf{Diffusion Models:}
% applications: Imag/video/audio generation
% Space: pixel space/latent space
% architecture: Unet-based or Transformer
Diffusion models have presented impressive success in various generative tasks including image synthesis, video generation, audio generation, etc. In general, diffusion models include a forward diffusion process for gradually corrupting real samples to learn a noise predictor and a reverse process to progressively generate less noisy samples via a trained denoising network. To reduce the computational complexity, latent diffusion models (LDMs) employ U-Net architectures to perform the diffusion on the latent space with lower feature dimensions. Recently, the diffusion transformer (DiT)~\cite{peebles2023scalable} has emerged as an effective replacement for the U-Net backbone in various diffusers like SiT~\cite{ma2024sit}, SD3~\cite{esser2024scaling}, VDT~\cite{lu2023vdt}, Latte~\cite{ma2024latte}, Sora~\cite{videoworldsimulators2024}, OpenSora~\cite{open-sora}, ViT-TTS~\cite{liu2023vit}, etc. However, existing transformer-based diffusion models concentrate on generating a single modality, constraining its potential application in multimodal generated content. Hence, our proposed AV-DiT is the first work to investigate how to utilize the DiT structure for joint audio and video generation.

\textbf{Joint Audio-Video Generation:}
% from single-modality generation to multimodal joint generation
% MM-Diffusion and Seeing and Hearing
Different from video generators that only create silent videos, joint audio-video generation aims to generate high-quality realistic videos that people can watch and listen to at the same time. MM-Diffusion~\cite{ruan2023mm} is the pioneering work to adopt diffusion models for generating audio-video pairs. More specifically, MM-Diffusion consists of separate video and audio branches to perform the joint multimodal denoising, where the random-shift based attention module is proposed to learn the consistency between audio and video modalities. Furthermore, Seeing and Hearing~\cite{xing2024seeing} proposes a multimodal latent aligner based on ImageBind~\cite{girdhar2023imagebind} to align the well-learnt latent features from pre-trained audio and video diffusion models, achieving the cross-modal generation without training the model from scratch. Compared with existing joint audio-video generation models, our AV-DiT leverages a shared frozen diffusion model solely pre-trained on image data to generate the sounding videos via introducing lightweight trainable adapters.

\textbf{Parameter-efficient Generative Models:}
% 1. Simply introduce parameter-efficient fine-tuning technologies: adapter, lora, and prompt, and their advantages
% 2. Some works that adopt PEFT into generative models: such as "SimDA: Simple Diffusion Adapter for Efficient Video Generation", "I2V-Adapter: A General Image-to-Video Adapter for Diffusion Models"
Training or fine-tuning large diffusion models for specific downstream tasks is time-consuming and computationally expensive. Therefore, parameter-efficient generative models are proposed to adopt the parameter-efficient fine-tuning (PEFT) strategy to only tune partial layers while keeping the majority weights of pre-trained frozen. PEFT technologies mainly consist of adapter tunning for inserting bottleneck adapters~\cite{houlsby2019parameter}, prompt tuning for injecting learnable prompt tokens at input space~\cite{jia2022visual} and low-rank adaptation (LoRA) for approximating the model weights by a low-rank factorization~\cite{hu2021lora}. To sufficiently leverage the pre-trained image generator to produce video, some lightweight adapters or temporal layers are inserted into frozen LDMs to learn the temporal consistency among different video frames~\cite{xing2023simda, guo2023i2v, gong2024atomovideo}. Besides, the trainable LoRA layers can be injected into frozen pre-trained LDM to introduce controllable conditions, guiding the generation of personalized images without relying on full fine-tuning~\cite{yang2024lora}. Different from these works focusing on closed-domain adaptation (i.e. image-to-image or image-to-video), our AV-DiT adapts the frozen image generator to address the joint audio and video generation via lightweight trainable layers (i.e. LoRA or adapters).

%% file: sec/3_method.tex
\section{Method}

In this section, we present our proposed Audio-Visual Diffusion Transformer (AV-DiT) model for efficient joint audio and video generation. We first revisit the preliminary knowledge of vanilla diffusion transformer (DiT) on 2D image generation. Then, we introduce the problem definition of joint audio-video generation. Finally, we propose an efficient AV-DiT by leveraging the generalization ability of pre-trained DiT to simultaneously generate a video that can be watched and listened to.   

\subsection{Preliminaries}
\textbf{Revist Diffusion-based Generation:} Diffusion models commonly follow the paradigm that first transfers the given data into Gaussian noise by a forward diffusion process and then learns to reconstruct the data distribution via a reverse denoising process. Most existing diffusers are based on denoising diffusion probabilistic models (DDPMs)~\cite{ho2020denoising} that define a forward noising process to gradually corrupt the real sample $x_{0}$ from the data distribution $X$ with random Gaussian noise over a discrete time step $t$, which can be formulated as $q(x_t|x_{t-1}) = \mathcal{N}(x_t; \sqrt{1-\beta_t}x_{t-1}, \beta_t \bm{\mathrm{I}})$ where $(\beta_1, \beta_2, \dots, \beta_t, \dots, \beta_T)$ are noising schedule, and $T$ is the pre-defined number of total diffusion steps. Therefore, the forward process generates a series of corrupted samples $(x_1, x_2, \dots, x_t, \dots, x_T)$ with ascending levels of noise via Markovian chain. By the re-parameterization method, the $x_t$ can be directly obtained by $\sqrt{\bar{\alpha_t}}x_0 + \sqrt{1-\bar{\alpha_t}}\mathbf{\epsilon}$, where $\bar{\alpha_t} = \textstyle\prod_{i=1}^{t}\alpha_i$, $\alpha_t = 1 - \beta_t$, and $\epsilon\sim \mathcal{N}(\bm{0}, \bm{\mathrm{I}})$.

To recover the original data, the reverse process of diffusion models commonly leverages a learned denoising network $\theta$ to predict less noisy data $x_{t-1}$ from the noisy input $x_t$ at each timestep via inverting the forward process, which can be formulated as $p_{\theta}(x_{t-1}|x_t) = \mathcal{N}(x_{t-1}; \mu_\theta(x_t, t), {\sigma}_{t}^{2}\bm{\mathrm{I}})$. To update the parameters of such denoiser $\theta$, the training objective is to maximize a variational lower bound of the log-likelihood as $\mathcal{L}_{\theta} = \sum_{t} -p_{\theta}(x_0|x_1) + \mathcal{D}_{KL}(q(x_{t-1}|x_t, x_0)|| p_{\theta}(x_{t-1}|x_t))$, which can be simplified to minimize a mean squire loss between the denoising network prediction $\epsilon_{\theta}(x_t, t)$ and ground truth added noise in forward process, defined as follows: $\min_{\theta} \mathcal{L}_{simple} = \min_{\theta} \mathbb{E}_{x_t, t, \epsilon}{||\epsilon-\epsilon_{\theta}(x_t, t)||}_{2}^{2}$. Afterwards, the data can be generated by progressively sampling $x_{t-1}$ from $p_{\theta}(x_{t-1}|x_t)$ by a trained denoising network. 

\textbf{Vanilla Diffusion Transformer (DiT):}
A vanilla diffusion transformer (DiT) is proposed to replace the commonly used U-Net denoiser with a transformer structure ~\cite{vaswani2017attention}, which follows the design of latent diffusion models (LDMs) ~\cite{rombach2022high} to operate on the latent space for reducing the computational complexity of diffusion models in high-resolution pixel space. More specifically, the DiT first compresses the input image into the latent code $z$ with a lower feature dimension by an encoder $\mathcal{E}$ from the pre-trained variational autoencoder (VAE). Then, the transformer blocks consisting of multi-head self-attention (MHSA) and feed-forward network (FFN) modules are operated on the latent space to perform the forward diffusion and backward denoising processes, where the time embedding and class embedding are incorporated for conditioning the generation by adaptive normalization layers. Note that, the patch embedding layer and position encoding are used to transform the latent code into the input tokens of the transformer. Finally, the learned decoder $\mathcal{D}$ from VAE is adopted to recover the generated latent code $\hat{z}$ back to the generated image sample. In this work, we employ the DiT as the diffusion backbone and adapt this framework to address the joint generation of audio and video. More details can be found in the following sections. 

\textbf{Problem Definition:} 
Most existing diffusion models perform the forward and reverse process for generating single-modality output such as image, video, or audio. However, the objective of our work is to tackle a more challenging problem such as the joint generation of high-quality audio and video. To simplify the problem, we directly start from the reverse process to introduce how to generate audio and video modalities from the Gaussian noises having the same feature sizes as the ones of audio and video latent code. Given a paired noise $(z_T^a, z_T^v)$ randomly sampled from Gaussian distribution, the joint audio and video generation aims to train a joint denoising network $\theta_{av}$ to fit the reverse process by taking both modalities as the inputs and benefiting the generative quality of counterpart modality. In other words, the reverse process adopts the trained $\theta_{av}$ to predict less noisy video and less noisy audio by considering their own and counterpart modalities, which can be formulated as follows:
\begin{gather}
    p_{\theta_{av}}(z_{t-1}^a | (z_t^a, z_t^v)) = \mathcal{N}(z_{t-1}^a; \mu_{\theta_{av}}(z_t^a, z_t^v, t), {\sigma}_{t}^{2}\bm{\mathrm{I}}) \\
    p_{\theta_{av}}(z_{t-1}^v | (z_t^v, z_t^a)) = \mathcal{N}(z_{t-1}^v; \mu_{\theta_{av}}(z_t^v, z_t^a, t), {\sigma}_{t}^{2}\bm{\mathrm{I}})
\end{gather}
where $t=1, 2, \dots, T$ denotes the diffusion steps. The training objective of joint denoising network $\theta_{av}$ can be formulated as follows:
\begin{gather}
    \mathcal{L}_{\theta_{av}} = \mathbb{E}_{z_t^v, z_t^a, t, \epsilon_v, \epsilon_a}{||\epsilon_v-\epsilon_{\theta_{av}}(z_t^v, z_t^a, t)||}_{2}^{2} + {||\epsilon_a-\epsilon_{\theta_{av}}(z_t^a, z_t^v, t)||}_{2}^{2}
\end{gather}
where $\epsilon_v$ and $\epsilon_a$ mean the ground noises for corrupting video and audio modality in the forward process, respectively. 

\textbf{Motivation:}
In this paper, we construct our joint denoising network based on the diffusion transformer since its unified attention-based structure can be utilized by various modality domains such as video and audio. One naive method is to design a multimodal version of DiT and train it on the massive pairs of audio and video data from scratch, which consumes the expensive computation burden and data acquisition procedure. Thereby, we propose an audio-visual diffusion transformer (AV-DiT) for simultaneously generating audio and video modality, leveraging the shared DiT pre-trained on image data as well as lightweight modality-specific adapters. Inspired by the effectiveness of parameter-efficient fine-tuning on domain adaptation, we only optimize the newly introduced adapters while keeping the pre-trained DiT frozen to generalize its learnt knowledge to benefit audio and video joint generation. To achieve this objective, our AV-DiT aims to tackle three important challenges: 1. Enable pre-trained Image DiT for video generation; 2. Adapt the pre-trained image DiT for audio generation via mitigating the domain gap; 3. Generalize the pre-trained DiT for interacting or aligning features from audio and video modality. In the following content, the abbreviation `DiT' refers to the diffusion transformer pre-trained in image data unless otherwise specified.

\begin{figure*}[ht]
\centering
\vspace{-8mm}
\includegraphics[scale=0.5]{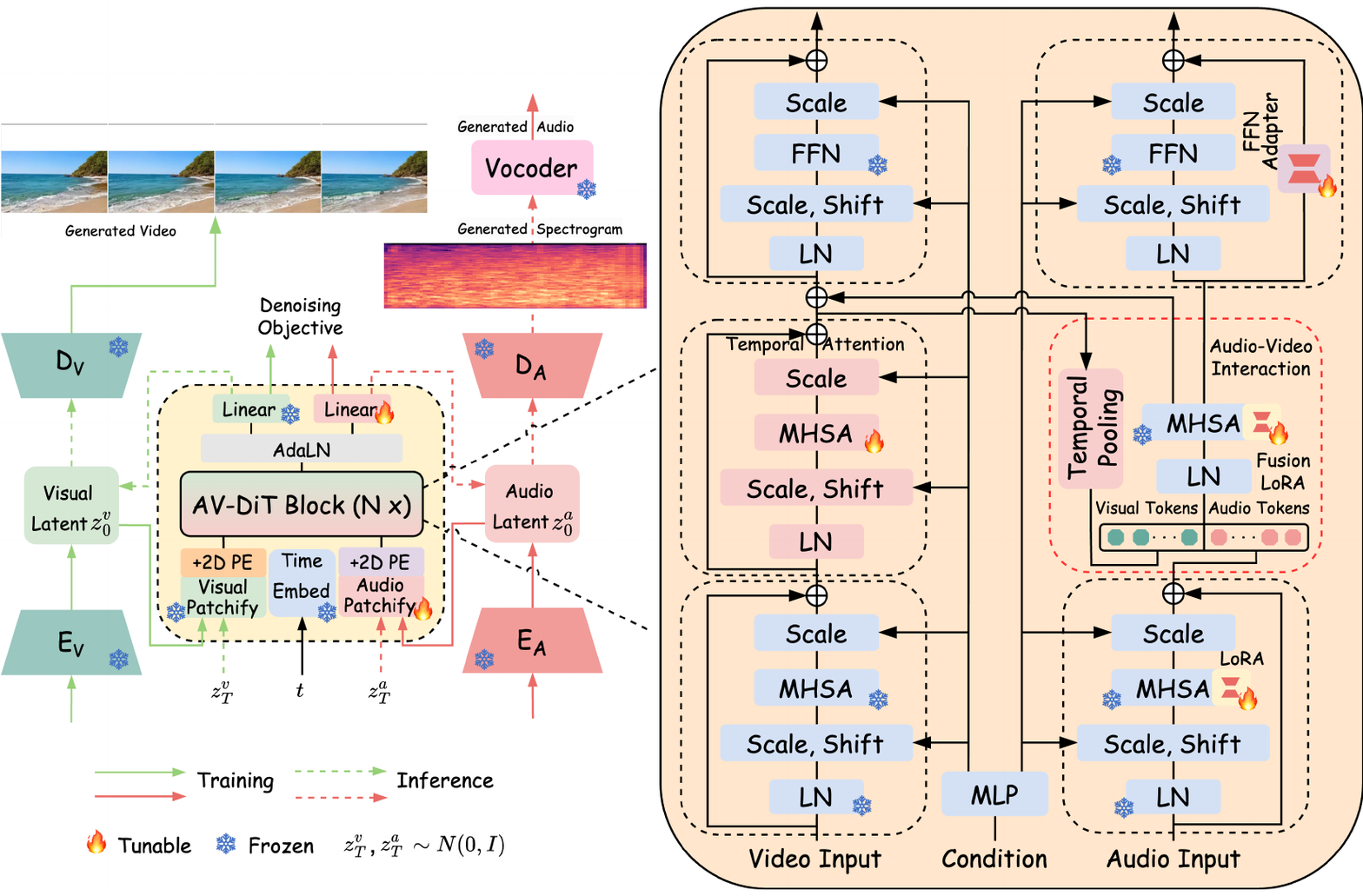}
\vspace{-11mm}
\caption{Illustration of our proposed AV-DiT for joint audio and video generation. Our AV-DiT leverages a shared frozen DiT backbone pre-trained on image-only data to simultaneously generate high-quality and realistic audio and video, where only inserted modality-specific adapters are trainable while the original pre-trained weights are frozen.}
%Note that, only added adapters are trainable while all other layers are frozen during training.
\label{figure 2}
\vspace{-7mm}
%\vspace{-1.5em}
\end{figure*}

\subsection{Audio-Visual Diffusion Transformer for Joint Generation}
To enable the denoising process in the latent space for joint audio and video generation, we propose an audio-visual diffusion transformer (AV-DiT) by leveraging the off-the-shell DiT backbone pre-trained on image-only data, aiming to make full use of its generalization ability to reduce the training cost and model complexity. As shown in Figure~\ref{figure 2},  our AV-DiT augments the shared image-based DiT with some modality-specific adapters to learn to recover the video and audio modality from Gaussian noise, where the original DiT weights are frozen while inserted parameters are trainable. 

\textbf{Video and Audio Latent Encoding:}
Our AV-DiT follows the LDMs to adopt the VAE encoder to project the input video and audio into the latent space before undergoing the AV-DiT denoising network. More specifically, the input videos $V\in \mathbb{R}^{B \times M \times H \times W \times 3}$ are first flattened along the batch $B$ and temporal dimension $F$, and are then extracted by a learned VAE encoder, resulting in the video latent features $z_v = \mathcal{E}_v(V)$, where $z_v 
 \in \mathbb{R}^{(B * M) \times \frac{H}{r_v} \times \frac{W}{r_v} \times c_v}$, $M$ means the number of video frames, $c_v$ is the video channel and $r_v>1$ denotes the video downsampling ratio. Meanwhile, the audio waveform is transformed into an image-like mel-spectrogram $A \in \mathbb{R}^{B \times T \times F}$ via the STFT operator which is then passed through a pre-trained VAE encoder to obtain the audio latent code $z_a = \mathcal{E}_a(A)$, where $z_a \in \mathbb{R}^{B \times \frac{T}{r_a} \times \frac{F}{r_a} \times c_a}$, $T$, $F$ and $c_a$ denote the temporal, frequency, and channel dimension, respectively, and $r_a > 1$ means the ratio of downsampling audio resolution. In our work, we employ the off-the-shell pre-trained VAEs from image LDM (i.e. Latent Stable Diffusion~\cite{rombach2022high}) and audio LDM (i.e. Tango~\cite{ghosal2023text}) to extract the video and audio latent codes respectively and save them locally to reduce the training memory. Furthermore, similar to the ViTs, the $z_v$ and $z_a$ are transformed into the sequence of tokens by respective patch embedding layers, which are then followed by the positional encoding to yield the video $x_v \in \mathbb{R}^{(B * M) \times L_v \times D}$ and audio $x_a \in \mathbb{R}^{B \times L_a \times D}$ inputs of our proposed multimodal diffusion transformer. In addition, the time information is incorporated into the diffusion transformer blocks via a time embedder. Note that, in our proposed diffusion transformer, we adopt the frozen visual patch embedding and time embedding layers from the pre-trained image-based DiT while employing a trainable patch embedder for specifically addressing the audio modality.     

\textbf{AV-DiT Block Design:}
Once the audio and video inputs are obtained, a sequence of proposed AV-DiT blocks is adopted to perform the joint diffusion and denoising processes. In each AV-DiT block, the video $x_v$ and audio $x_a$ inputs are normalized by a shared adaptive layer normalization (AdaLN) ~\cite{perez2018film} that regresses the scale $<\alpha_v^1, \alpha_a^1>$ and shift $<\beta_v^1, \beta_a^1>$ parameters from MLP conditioning block to introduce the time guidance. Then, the normalized video features undergo a frozen pre-trained multi-head self-attention (MHSA) to learn the spatial correlation within each video frame along $L_v$ dimension'. Meanwhile, audio features share the same frozen pre-trained MHSA since the learnt attention weights have been explored to be an effective initialization for the audio domain ~\cite{gong2021ast, lin2023vision}. However, to further alleviate the domain gap between audio and image modality, we inject trainable LoRA layers into projection modules of frozen MHSA to transfer the knowledge from the image into the audio domain. Afterwards, a pair of scaling parameters $<\gamma_v^1, \gamma_a^1>$ from the conditioning block is used to control the information flow prior to the residual connection. It is worth noting that the weights of the conditioning block are inherited from the one of frozen DiT to provide consistent information guidance for both audio and video diffusion.  

To compensate for the shortage of DiT in modelling temporal dependency, a new temporal adapter is inserted after the frozen spatial MHSA to maintain the temporal consistency, where the new adapter has the same structure as the frozen MHSA but it is trainable. More specifically, the output video features from frozen MHSA are permuted into $x_{v, f} \in \mathbb{R}^{(B * L_v) \times M \times D}$ by swapping the $M$ and $L_v$ dimensions, which are then passed through the temporal adapter to learn the temporal dependency among different video frames. To reduce the complexity, the temporal adapter enables the feature compression on the query and key projector of the MSAH block, whose compression ratio will be explored in the following section. Moreover, the factors regressed from the frozen contion block are used to introduce the time information into temporal adapter via adaptive layernorm.

To bridge audio and video branches and learn the multimodal alignment for better joint generation, inspired by SD3~\cite{esser2024scaling}, the video tokens are first pooled through temporal dimension and then concatenated with audio tokens to be fed into a MHSA for mutual interaction between two modalities. Instead of~\cite{esser2024scaling} using a trainable MHSA, our AV-DiT reuses a frozen MHSA block from DiT and augments it with LoRA layers to adapt the learning from close-domain knowledge to audio-visual interaction,  generating the audio-steered video features and video-steered audio features to refine the video and audio tokens, respectively. 

In the following, refined video and audio tokens are processed by a shared frozen feed-forward network (FFN) from pre-trained DiT for global extraction of each modality, where the time conditions are injected into each branch via adaLN as previously operated.  Besides, a simple learnable adapter consisting of bottleneck MLP is parallelly connected with FFN of audio branch, adjusting the feature distribution to further adapt the image knowledge benefit from modeling audio modality. Eventually, the output video and audio features of AV-DiT block are produced to again pass through next block for iterative processing.   

\textbf{Video and Audio Latent Decoding:}
After the final AV-DiT block, the sequences of video and audio tokens are required to be decoded into the predicted noises and diagonal covariance of corresponding modalities. To do so, a shared AdaLN followed by two separate liner decoding layers are adopted to decode video and audio tokens into corresponding predicted latent codes respectively, generating the original spatial feature layout via the feature rearrangement to obtain the predicted noise and covariance of video and audio for diffusion objective. Note that, the decoding layer from the video is directly from frozen DiT, while the other one is initialized and trained from scratch. Once diffusion forward is finished, two sampled Gaussian noises are separately fed into the trained AV-DiT to progressively perform noise removal, yielding the less noisy audio and video latent code at the last diffusion time step. Next, the decoders from pre-trained VAEs from image LDM and audio LDM are employed to simultaneously reconstruct the video and audio latent features back generated video frames and audio mel-spectrogram. Last but not least, a pre-trained HiFi-GAN~\cite{kong2020hifi} is used as a vocoder to transform the audio mel-spectrograms into the audio waveforms.   

%% file: sec/4_experiments.tex
\section{Experiments}
\subsection{Experimental Setup}
\label{sec:exp_setup}

\textbf{Datasets:}
Following previous work~\cite{ruan2023mm}, we evaluate our proposed AV-DiT on two high-quality datasets including Landscape~\cite{lee2022sound} and AIST++~\cite{li2021ai} for joint audio and video generation. \textbf{Landscape} is a high-fidelity dataset including video and audio streams and it features nine diverse nature scenes including raining, splashing water, thunder, underwater burbling, etc. In addition, the Landscape dataset contains 928 videos crawled from YouTube as listed in \cite{lee2022sound} and creates 1,000 video clips of 10 seconds without overlap, spanning approximately 2.7 hours in total. \textbf{AIST++} is acquired from the AIST dataset~\cite{tsuchida2019aist}, featuring street dancing videos with 60 copyright-cleared songs. The dataset consists of 1,020 video clips having a total duration of 5.2 hours. Moreover, the resolution of video frames is cropped upon $1024\times1024$ from the center of the raw videos to present clear characters. 

\textbf{Implementation:}
% preprocess, model architecture, diffusion configuration
For data pre-processing, following the work~\cite{ruan2023mm}, 16 video frames are sampled to construct a video clip and are then cropped into $256\times256$ resolution. Based on the duration of each video clip, we crop out the corresponding audio signal and then truncate or pad it into a 1.6-second waveform at the sampling rate of 16 KHz. For the diffusion stage, we adopt the off-the-shelf pre-trained VAEs from Stable Diffusion to transform the input video into latent code with the size of $(32\times32\times4)$. Meanwhile, the STFT operator and pre-trained VAE from audio LDM (i.e. Tango~\cite{ghosal2023text}) are used to project the input audio into a latent feature with size of $(40\times16\times8)$. Our AV-DiT adopts the frozen pre-trained DiT XL/2 backbone as well as lightweight trainable layers to predict the noise during the forward process. Moreover, the used DiT is pre-trained on the ImageNet with the resolution of $256\times256$ by using 28 transformer layers with 16 attention heads and a feature dimension of 1152. We retain the diffusion configuration as ADM~\cite{dhariwal2021diffusion} which uses a linear noise schedule ranging from $1\times10^{-4}$ to $2\times10^{-2}$ via 1,000 time steps. In addition, we train our proposed AV-DiT for 100K iterations with a batch size of 16 and a constant learning rate of $5\times 10^{-4}$ via AdamW~\cite{loshchilov2017decoupled} optimizer. Note that, only newly inserted layers (e.g., adapters, LoRA, audio patch embedding and decoding layers), and bias terms are trainable, while the pre-trained DiT backbone is kept frozen during the training.  

\begin{figure*}[t]
%\centering
\includegraphics[scale=0.89]{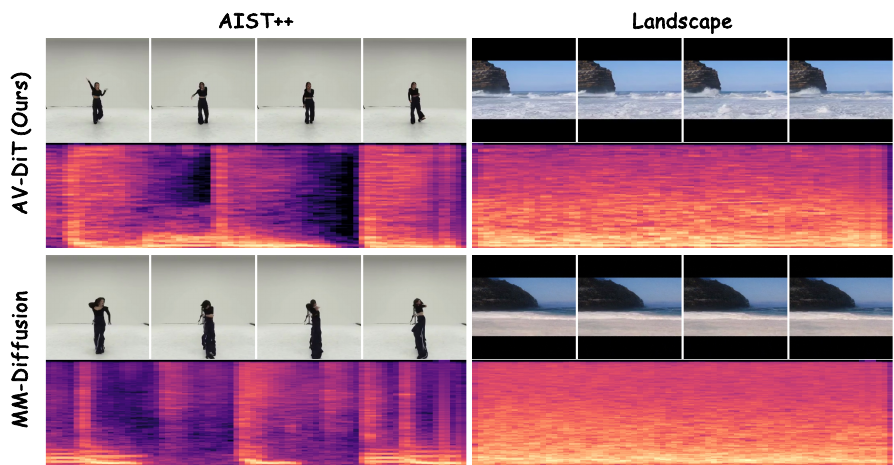}
%\vspace{-7.5em}
\caption{Qualitative examples of our AV-DiT and the MM-Diffusion model. Compared with MM-Difusion, our method generates higher quality and more realistic videos. Meanwhile, our generated audio spectrogram involves fewer artifacts and restores more approximate structures reflecting the visual scenes. For example, our generated audio sample of Landscape scenes possesses more details that demonstrate the sound of waves lapping on the shore.}
\vspace{-5mm}

\label{figure 3}
%\vspace{-1.5em}
\end{figure*}

\textbf{Evaluation:}
% inference + objective evaluation metrics
To evaluate the quality of generated videos, following previous works~\cite{ruan2023mm, yu2022generating, ge2022long}, we adopt the Frechet Video Distance (FVD) and Kernel Video Distance (KVD) by implementing the I3D~\cite{carreira2017quo} video classifier pre-trained on Kinetics-400~\cite{carreira2017quo}. For audio evaluation, we calculate the Frechet audio distance (FAD)~\cite{ruan2023mm} between the pairs of ground-truth and generated audio in the space of latent features extracted by AudioCLIP~\cite{guzhov2022audioclip}. As for all metrics, the lower evaluation score represents a better quality. In our experiments, we randomly generate 2,048 samples by using our trained AV-DiT to calculate objective evaluation scores, where the original generated and real videos are cropped out $64 \times 64$ resolution for computation efficiency. To have a fair comparison, we evaluate all experiments by averaging 5 runs to reduce randomness. 

\subsection{Comparison with SOTA methods}
\begin{figure*}[t]
\centering
\includegraphics[scale=0.55]{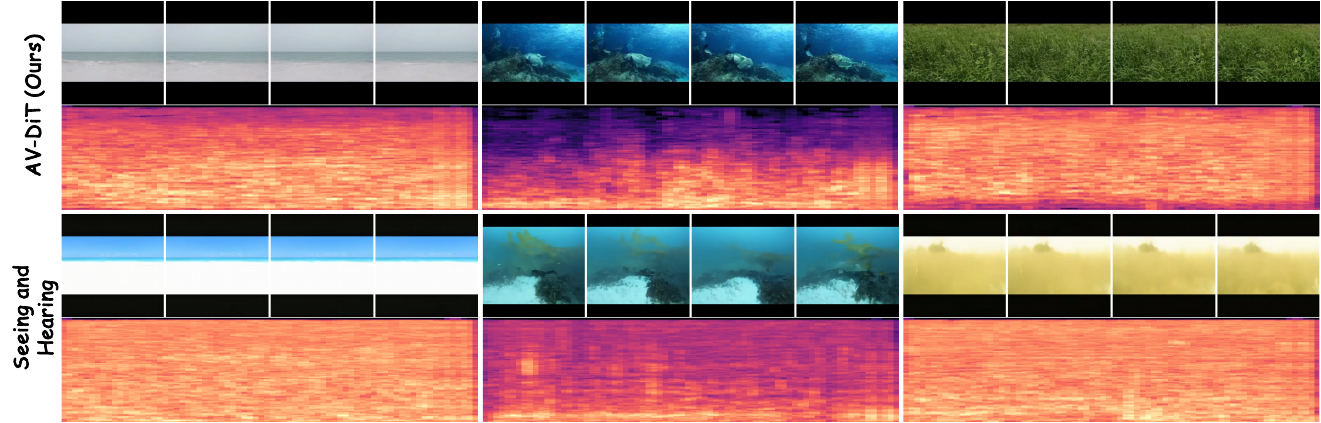}
%\vspace{-1.0em}
\caption{Generatation results on Landscape. Our AV-DiT yields higher quality and more realistic sounding videos than Seeing and Hearing~\cite{xing2024seeing}.}
\label{figure 4}
\vspace{-5mm}
\end{figure*}

% main tables for landscape and aist
As shown in Table~\ref{table 1}, we compare our proposed AV-DiT with existing state-of-the-art methods (i.e. DIGAN~\cite{yu2022generating}, TATS~\cite{ge2022long}, MM-Diffusion~\cite{ruan2023mm}, Seeing and Hearing~\cite{xing2024seeing}) on AIST++ and Landscape datasets. In general, our AV-DiT achieves competitive or even superior performance than existing methods on joint audio and video generation while involving significantly reduced trainable parameters. First, as for AIST++ dataset, compared with existing baselines, our AV-DiT achieves the best results in FVD ($68.88$), presenting impressive performance in generating high-quality videos as shown in Figure~\ref{figure 3}. Although recent MM-Diffusion obtained a lower KVD score than our AV-DiT, it adopts the full training of audio and video branches and involves much more trainable parameters than ours ($159.91$M vs $426.16$M). In addition, MM-Diffusion uses a two-stage generation that first produces a small resolution and then upscales to a large resolution by an extra pre-trained super-resolution network, while our AV-DiT directly generates the target resolution in an end-to-end manner due to our efficient design.  For audio evaluation, our AV-DiT yields a superior FAD score to existing methods without any audio-specific training, demonstrating that our AV-DiT efficiently adapts the pre-trained image diffusion to generate high-fidelity audio. Second, as for the Landscape dataset, our AV-DiT also achieves competitive performance in all evaluation metrics including FVD ($172.69$), KVD ($15.41$), and FAD ($11.17$) with reduced tunable parameters. It is worth mentioning that we generated the same number of videos (200 samples) as Seeing and Hearing for a fair comparison on Landscape as seen in Figure~\ref{figure 4}. As shown in Table~\ref{table 1}, we find that our AV-DiT attains better evaluation scores than Seeing and Hearing in all objective metrics, presenting excellent generative performance of sounding videos. Finally, we compare the performance of our methods and the MM-Diffusion baseline in inference efficiency. From Table~\ref{table 1}, we observe that our AV-DiT possesses a three times faster inference speed than MM-Diffusion, showing the efficient generation capability of joint audio and video.               

\begin{table}[t]
\setlength{\abovecaptionskip}{0.2cm}
\renewcommand\arraystretch{1.2}
\footnotesize
%\small
%\scriptsize
\centering
%\captionsetup{font={scriptsize}}
\caption{Comparison between our AV-DiT and existing SOTA methods.  Since Seeing and Hearing~\cite{xing2024seeing} only evaluates 200 samples, our AV-DiT also uses the same number of samples for fair comparison. For evaluation, we used the same scripts as MM-Diffusion~\cite{ruan2023mm}. Note that $*$ means the reproduced results by using the released model weights from the authors.}
\label{table 1}
\resizebox{\linewidth}{!}{
\begin{tabular}{c|ccc|ccc|c|c}
\hline
\multirow{2}{*}{\textbf{Model}} & \multicolumn{3}{c|}{\textbf{AIST++}} & \multicolumn{3}{c|}{\textbf{Landscape}} & \multirow{2}{*}{\textbf{Param.} $\downarrow$} & \multirow{2}{*}{\begin{tabular}[c]{@{}c@{}}\textbf{Inference} \\ \textbf{Speed} $\uparrow$\end{tabular}} \\ \cline{2-7}
                       & FVD $\downarrow$     & KVD $\downarrow$     & FAD $\downarrow$     & FVD $\downarrow$      & KVD $\downarrow$      & FAD $\downarrow$      &                             &                                 \\ \hline
GroundTrugh           & 8.73         &0.0036         &8.46
         &17.83          &-0.12          &7.51          & -                            & -                                \\
DIGAN (ICLR 2022)~\cite{yu2022generating}                  &119.47
         &35.84         & -        & 305.36         & 19.56         & -          & -                            & -                               \\
TATS-base (ECCV 2022)~\cite{ge2022long}             &267.24         &41.64
         & -        &267.24          & 41.64         & -         & -                            & -                                \\
MM-Diffusion* (CVPR 2023)~\cite{ruan2023mm}          &98.69         &\textbf{18.90}
         &10.58         & 186.09         & \textbf{9.21}         &\textbf{10.61}          & 426.16M                            & 0.009 sample/sec                                 \\

\textbf{AV-DiT (Ours)}          & \textbf{68.88}        & 21.01        & \textbf{10.17}       &\textbf{172.69}          &15.41          &11.17          & \textbf{159.91M}                             & \textbf{0.032} sample/sec                                \\ \hline
Seeing and Hearing\footnotemark[1] (CVPR 2024)~\cite{xing2024seeing}     & -        & -        & -        & 326.23         &9.20          &  \textbf{12.76}        & -                            &-                                 \\
\textbf{AV-DiT (Ours)}, 200 samples          & -        &-         &-        &\textbf{260.50}          &\textbf{9.15}          &14.15          &  \textbf{159.91M }                          & \textbf{0.032} sample/sec                                 \\ \hline
\end{tabular}
}
\vspace{-5mm}
\end{table}
\footnotetext[1]{We appreciate the authors of Seeing and Hearing~\cite{xing2024seeing} providing us their generated samples on Landscape.}

\subsection{Ablation Study}

\begin{minipage}{\textwidth}
\scriptsize
\begin{minipage}[t]{0.48\textwidth}
\setlength{\abovecaptionskip}{0.2cm}
\renewcommand\arraystretch{1.33}
\centering
\makeatletter\def\@captype{table}
\caption{Influence of various adapter layers}
\label{table 2}
\begin{tabular}{c|ccc}
\hline
\textbf{Model}                      & \textbf{FVD} $\downarrow$ & \textbf{KVD} $\downarrow$ & \textbf{FAD} $\downarrow$ \\ \hline
AV-DiT                     & \textbf{68.88}    & 21.01    &\textbf{10.17 }    \\ \hline
w/o Video temporal adapter &365.71     &101.33     &10.23     \\ \hline
w/o Audio FFN adapter      &72.81     &22.03     & 10.28    \\
w/o Audio LoRA             & 69.46    & 20.80    &10.22     \\
w/o Audio LoRA and adapter & 74.01    & 21.63    & 10.21    \\ \hline
w/o Fusion                 & 72.11     & \textbf{20.39}    &11.87     \\
w/o Fusion LoRA            &73.62     &21.95     &10.19     \\ \hline
\end{tabular}
\end{minipage}
\begin{minipage}[t]{0.48\textwidth}
\setlength{\abovecaptionskip}{0.2cm}
\renewcommand\arraystretch{1.2}
\centering
\makeatletter\def\@captype{table}
\caption{Different adapter ratios}
\label{table 3}
\begin{tabular}{ccc|ccc}
\hline
\begin{tabular}[c]{@{}c@{}}\textbf{Video}\\ \textbf{Ratio}\end{tabular} & \begin{tabular}[c]{@{}c@{}}\textbf{Audio}\\ \textbf{Ratio}\end{tabular} & \begin{tabular}[c]{@{}c@{}}\textbf{Fusion}\\ \textbf{Ratio}\end{tabular} & \textbf{FVD} $\downarrow$ & \textbf{KVD} $\downarrow$ & \textbf{FAD} $\downarrow$ \\ \hline
8                                                     & 2                                                     & 2                                                      & 68.88    & 21.01    &10.17     \\
4                                                     & 2                                                     & 2                                                      &67.60     &22.68     &11.06     \\
2                                                     & 2                                                     & 2                                                      &\textbf{67.13}     & 22.68    &10.11     \\
4                                                     & 4                                                     & 2                                                      &72.59     &23.48     &10.15     \\
4                                                     & 8                                                     & 2                                                      & 70.51    &\textbf{20.70}     &10.18     \\
4                                                     & 2                                                     & 4                                                      &72.71     &21.06     &10.25     \\
4                                                     & 2                                                     & 8                                                      &70.10     &20.95     &\textbf{10.03}     \\ \hline
\end{tabular}
\end{minipage}
\end{minipage}

\textbf{Influence of Various Adapter Layers:}
To verify the effects of different adapter layers including LoRA, temporal adapter, and FFN adapter on generative performance, we establish various reference models by removing corresponding layers as shown in Table~\ref{table 2}. First, omitting the temporal adapter leads to an obvious performance drop in terms of FVD (from $68.88$ to $365.71$) and KVD (from $21.01$ to $101.33$), showing that keeping the temporal consistency is very crucial for video generation. Second, the reference models without either an FFN adapter or LoRA in the audio branch attain a worse performance than our AV-DiT, presenting that audio-specific adapters are useful for the frozen pre-trained image generator to extend to joint audio and video generation. Finally, when removing the LoRA from pre-trained MHSA in the fusion module, the performance will be decreased compared with AV-DiT, demonstrating that directly adapting the self-attention of the frozen DiT makes it hard to perform the multimodal interaction.  

\textbf{Different Adapter Ratio:}
To reduce the model complexity and computing burden, the downsampling operation is applied in each adapter layer. Therefore, we conduct the ablation study to analyze the influence of the choice of adapter ratios on generating performance as shown in Table~\ref{table 3}. We find that increasing the ratio of the video temporal adapter will lead to a performance drop in terms of FVD, showing that properly increasing the trainable parameters of the temporal adapter is beneficial for learning the temporal dependency towards better generated videos. Moreover, it is worth mentioning that the best FAD score is achieved by setting a small ratio of audio adapters and a large ratio of fusion adapters. By considering the trade-off between performance and computing complexity, we determine $(8, 2, 2)$ as the ratio for video, audio and fusion part, respectively.   

\textbf{Scaling Backbone:}
When we replace the frozen DiT backbone of AV-DiT with a pre-trained one on large image resolution (i.e. $512\times512$), we can observe improved performance in Table~\ref{table 4}. That means our parameter-efficient AV-DiT can benefit from more powerful pre-trained image generators.   

\textbf{Cross-attention Fusion:}
Our AV-DiT adopts a joint self-attention for connecting the audio and video features for multimodal alignment. To explore the efficiency of our audio-video fusion mechanism, we also design the other reference model using cross-attention based fusion. More specifically, in each AV-DiT block, two cross-attention blocks are separately inserted before the FFN module of each branch for bi-directional conditioning audio and video. As shown in Table~\ref{table 5}, we can find that our AV-DiT with self-attention fusion outperforms better than the one with cross-attention fusion while involving lower model complexity and trainable parameters.

\begin{minipage}{\textwidth}
\scriptsize
\begin{minipage}[t]{0.45\textwidth}
\setlength{\abovecaptionskip}{0.58cm}
\renewcommand\arraystretch{1.33}
\centering
\makeatletter\def\@captype{table}
\caption{Different DiT backbones}
\label{table 4}
\begin{tabular}{c|ccc}
\hline
\textbf{Model}                                                                   & \textbf{FVD} $\downarrow$ & \textbf{KVD} $\downarrow$ & \textbf{FAD} $\downarrow$ \\ \hline
\begin{tabular}[c]{@{}c@{}}AV-DiT \\ (256*256 Backbone)\end{tabular}      & 68.88    & 21.01    &\textbf{10.17}     \\ \hline
\begin{tabular}[c]{@{}c@{}}AV-DiT (ours)\\ (512*512 Backbone)\end{tabular} & \textbf{67.26 }   & \textbf{20.02}    & 10.23    \\ \hline
\end{tabular}
\end{minipage}
\begin{minipage}[t]{0.45\textwidth}
\setlength{\abovecaptionskip}{0.2cm}
\renewcommand\arraystretch{1.33}
\centering
\makeatletter\def\@captype{table}
\caption{Self-attention fusion V.S. Cross-attention fusion}
\label{table 5}
\begin{tabular}{c|cccc}
\hline
\textbf{Model}                                                                          & \textbf{FVD} $\downarrow$ & \textbf{KVD} $\downarrow$ & \textbf{FAD} $\downarrow$ & \textbf{Param.}\\ \hline
\begin{tabular}[c]{@{}c@{}}AV-DiT (ours)\\ (Self-attention) Fusion\end{tabular}  & \textbf{68.88}    & 21.01    &\textbf{10.17}  & \textbf{159.91M }  \\ \hline
\begin{tabular}[c]{@{}c@{}}AV-DiT (ours)\\ (Cross-attention) Fusion\end{tabular} &71.57     &\textbf{20.19}     &11.43  & 289.82M   \\ \hline
\end{tabular}
\end{minipage}
\end{minipage}

% \subsection{Limitations}
% \label{sec:limitations}
% Our experiments mainly focus on unconditional audio and video generation. It is meaningful to study the effectiveness of the proposed AV-DiT in class-conditional and text-conditional generation of audio and video. Notably, our AV-DiT can be flexibly extended into the conditional joint generation of audio and video via injecting extra text prompts or audio-visual captions with minimal adjustment. Although our AV-DiT is more efficient than MM-Diffusion, it is required to be further improved for deployment in real-time applications. In future work, we plan to improve the generative efficiency of joint audio and video by accelerating the inference speed.     

%% file: sec/5_conclusion.tex
\section{Conclusion}
\label{sec:conclusion}
In this work, we proposed an AV-DiT, the first multimodal diffusion transformer architecture for efficient joint audio and video generation. Our AV-DiT leverages a shared frozen DiT backbone pre-trained on image-only data and lightweight trainable layers to jointly generate audio and video. To do so, our AV-DiT equips the frozen image generator with the ability to tackle the temporal consistency for video generation, mitigate the domain gap for audio generation, and interact with audio and video for multimodal alignment. Extensive experiments on benchmark datasets present that our AV-DiT achieves competitive or even better generative performance than state-of-the-art methods while having significantly reduced trainable parameters and more efficient inference ability.

\textbf{Limitations.} Our experiments mainly focus on unconditional audio and video generation. It is meaningful to study the effectiveness of the proposed AV-DiT in class-conditional and text-conditional generation of audio and video. Notably, our AV-DiT can be flexibly extended into the conditional joint generation of audio and video via injecting extra text prompts or audio-visual captions with minimal adjustment. Although our AV-DiT is more efficient than MM-Diffusion, it is required to be further improved for deployment in real-time applications. In future work, we plan to further improve the generative efficiency of joint audio and video by accelerating the inference speed.     

\textbf{Broader Impact.} Our AV-DiT model could be applied to enhance education, entertainment, and accessibility by generating synchronized audio-visual content, while also posing ethical concerns regarding the potential misuse of synthetic media.

%% file: sec/appendix.tex
\appendix

% \section{Appendix / supplemental material}

\section{Supplemental Material}

% Optionally include supplemental material (complete proofs, additional experiments and plots) in appendix.
% All such materials \textbf{SHOULD be included in the main submission.}

This supplemental material provides
    more implementation details including architecture and training process. In addition, we exhibit more generated sounding video examples to show the effectiveness of our AV-DiT. To have a better watching and listening experience,  please feel free to check our demo videos with audio available in index.html.

\subsection{More Implementation Details}

For detailed infromation of our implementation, see Table~\ref{table 6}.

\begin{table}[ht]
\setlength{\abovecaptionskip}{0.2cm}
\renewcommand\arraystretch{1.2}
\footnotesize
%\small
%\scriptsize
\centering
%\captionsetup{font={scriptsize}}
\caption{Comparison with SOTA methods}
\label{table 6}
\begin{tabular}{cc}
\hline
\multicolumn{1}{c|}{\textbf{Model Configuration}}                    & \textbf{Audio-Visual Diffusion Transformer (AV-DiT)} \\ \hline
\multicolumn{1}{l|}{\textbf{Architecture}}                           &                                                      \\
\multicolumn{1}{c|}{Feature Dimension}                               & 1152                                                 \\
\multicolumn{1}{c|}{Attention Head}                                  & 16                                                   \\
\multicolumn{1}{c|}{Transformer Depth}                               & 28                                                   \\
\multicolumn{1}{c|}{Compression Ratio in Video Temporal Adapter}     & 8                                                    \\
\multicolumn{1}{c|}{Compression Ratio in Audio LoRA and FFN Adapter} & 2                                                    \\
\multicolumn{1}{c|}{Compression Ratio in Fusion LoRA}                & 2                                                    \\ \hline
\multicolumn{1}{l|}{\textbf{Diffusion Process}}                      &                                                      \\
\multicolumn{1}{c|}{Diffision Steps}                                 & 1000                                                 \\
\multicolumn{1}{c|}{Diffusion Noise Scheduler}                       & Linear                                               \\
\multicolumn{1}{c|}{Prediction Objective}                            & Noise Prediction                                     \\
\multicolumn{1}{c|}{Sampling Method}                                 & DDPM                                                 \\
\multicolumn{1}{c|}{Sampling Steps}                                  & 250                                                  \\ \hline
\multicolumn{1}{l|}{\textbf{Input Data}}                             & \multicolumn{1}{l}{}                                 \\
\multicolumn{1}{c|}{Video Shape}                                     &   $16 \times 256 \times 256 \times 3$                               \\
\multicolumn{1}{c|}{Video FPS}                                       & 10                                                   \\
\multicolumn{1}{c|}{Audio Spectrogram Shape}                         & $160 \times 64$                                             \\
\multicolumn{1}{c|}{Audio Sample Rate}                               & 16K                                                  \\ \hline
\multicolumn{1}{l|}{\textbf{Training Setting}}                       & \multicolumn{1}{l}{}                                 \\
\multicolumn{1}{c|}{Learning Rate}                                   & 5e-4                                                 \\
\multicolumn{1}{c|}{Optimizer}                                       & AdamW                                                \\
\multicolumn{1}{c|}{Batch Size}                                      & 16                                                   \\
Loss Function                                                        & MSE                                                  \\
\multicolumn{1}{c|}{Training Iteations}                              & 100k                                                 \\
\multicolumn{1}{c|}{Training Hardware}                               & NVIDIA RTX $A6000$ GPU                                      \\ \hline
\end{tabular}
\vspace{-1.0em}
\end{table}

\subsection{Additional Qualitative 
 Examples}

\begin{figure*}[t]
\centering
\includegraphics[width=\linewidth]{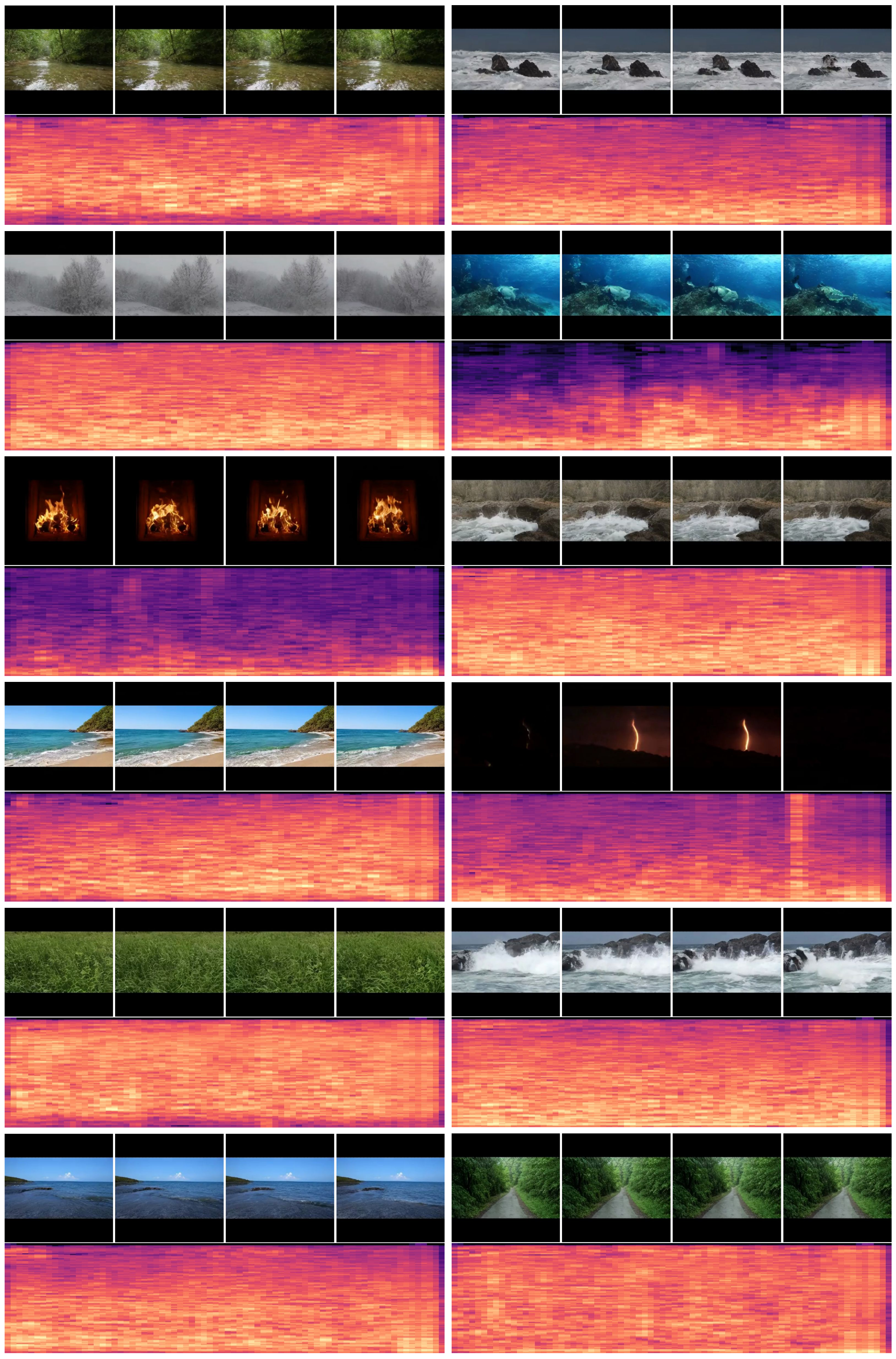}
%\vspace{-1.0em}
\caption{More results from our AV-DiT model on Landscape dataset. The generated audio and video are consistent with each other.}
%Note that, only added adapters are trainable while all other layers are frozen during training.
\label{figure 5}
%\vspace{-1.5em}
\end{figure*}

\begin{figure*}[t]
\centering
\includegraphics[width=\linewidth]{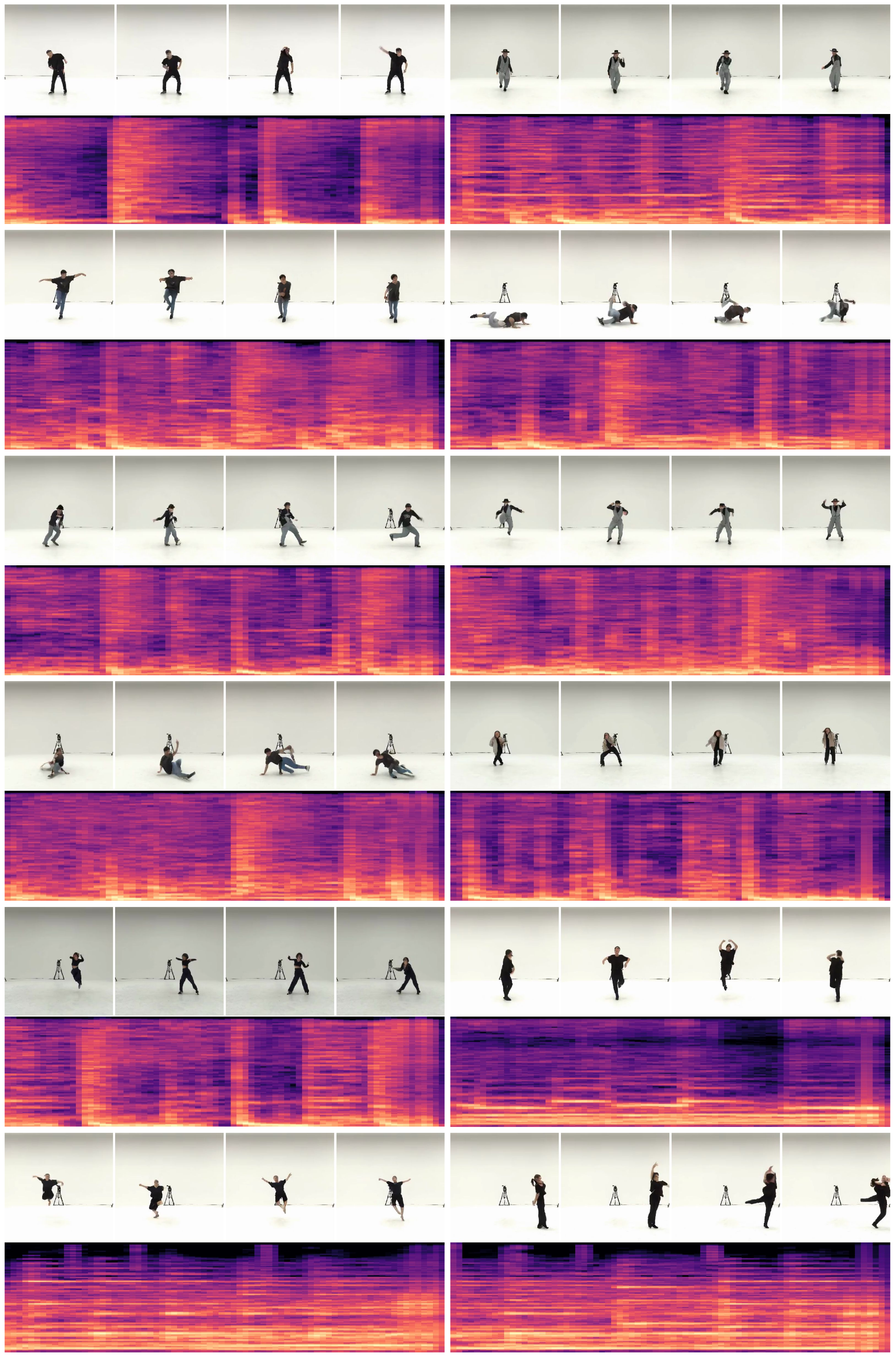}
%\vspace{-1.0em}
\caption{More results from our AV-DiT model on AIST++ dataset. The dance movements and music beats are well aligned.}
%Note that, only added adapters are trainable while all other layers are frozen during training.
\label{figure 6}
%\vspace{-1.5em}
\end{figure*}

For additional results on Landscapes, please check Figure~\ref{figure 5}.

For more results on AIST++, please refer to Figure~\ref{figure 6}.